\documentclass[%
 aip,
 amsmath,amssymb,
 reprint,%
]{revtex4-1}

\usepackage{graphicx}
\usepackage{dcolumn}
\usepackage{bm}

\usepackage[utf8]{inputenc}
\usepackage{mathptmx}
\usepackage{etoolbox}

\makeatletter
\def\@email#1#2{%
 \endgroup
 \patchcmd{\titleblock@produce}
  {\frontmatter@RRAPformat}
  {\frontmatter@RRAPformat{\produce@RRAP{*#1\href{mailto:#2}{#2}}}\frontmatter@RRAPformat}
  {}{}
}%
\makeatother
\begin{document}

\preprint{AIP/123-QED}

\title[Internal noise in deep neural networks]{Internal noise in deep neural networks: interplay of depth, neuron number, and noise injection step.}
\author{D.A. Maksimov}%
\author{V.M. Moskvitin}%
\author{N. Semenova}%
 \email{semenovani@sgu.ru}
 \affiliation{Saratov State University, Astrakhanskaya str. 83, Saratov 410012, Russia}%

\date{\today}

\begin{abstract}
This paper examines the influence of internal Gaussian noise on the performance of deep feedforward neural networks, focusing on the role of the noise injection stage relative to the activation function. Two scenarios are analyzed: noise introduced before and after the activation function, for both additive and multiplicative noise influence. The case of noise before activation function is similar to perturbations in the input channel of neuron, while the noise introduced after activation function is analogous to noise occurring either within the neuron itself or in its output channel. The types of noise and the method of their introduction were inspired by analog neural networks.
 The results show that the activation function acts as an effective nonlinear filter of noise. Networks with noise introduced before the activation function consistently achieve higher accuracy than those with noise applied after it, with additive noise being more effectively suppressed in this case. For noise introduced after the activation function, multiplicative noise is less detrimental than additive noise, and earlier hidden layers contribute more significantly to performance degradation due to cumulative noise amplification governed by the statistical properties of subsequent weight matrices.
 The study also demonstrates that pooling-based noise reduction is effective in both cases when noise is introduced before and after the activation function, consistently improving network performance.
\end{abstract}

\maketitle

\begin{quotation}
Over the past few years, artificial neural networks (ANNs) have found their application in solving many problems. In terms of computation, ANN modeling is a very resource-intensive task. Despite the existence of high-power computing clusters with the ability to parallelize computations, modeling a neural network on digital equipment is a bottleneck in network scaling, speed of receiving or processing information and energy efficiency. In recent years, more and more researchers in the field of neural networks are interested in creating hardware networks in which neurons and the connection between them represent a real device capable of learning and solving problems. However, experimental and hardware setups always contain noise of various natures, so studying their influence is a pressing problem, the solution of which will help improve the efficiency of training in the presence of noise. The study of the influence of various types of noise within the framework of machine learning is aimed at determining the properties of noise effects that can be accumulated by a neural network, or, conversely, be suppressed by a network itself.
\end{quotation}

\section{Introduction}\label{sec:intro}
Over the past several years, artificial neural networks (ANNs) have been widely applied to a variety of tasks, ranging from pattern recognition to the prediction of complex behaviour.

From a computational perspective, the simulation of artificial neural networks is a highly resource-intensive task. As neural networks and the problems they address continue to grow in complexity, we may soon approach a computational bottleneck \cite{Hasler2013,Gupta2015}, where the demands of these tasks exceed the capabilities of modern computers and high-performance computing clusters. Despite the availability of powerful parallel computing systems, the digital simulation of neural networks remains a limiting factor in terms of scalability, processing speed, and energy efficiency.
In recent years, increasing attention has been devoted to the development of hardware-based neural networks \cite{Karniadakis2021} (often referred to as analog neural networks in the literature), in which neurons and their interconnections are implemented as physical devices capable of learning and performing computational tasks. This approach involves the physical realization of both neurons and synaptic connections, enabling significant improvements in computational speed and energy efficiency \cite{Aguirre2024,Chen2023}.

Within this paradigm, neural networks are not simulated on conventional digital hardware but are realized as physical systems capable of learning and performing computational tasks. The neurons and their interconnections are implemented at the hardware level, meaning that the network operates according to underlying physical principles rather than numerical simulation. In recent years, there has been an exponential growth in research devoted to hardware implementations of ANNs. Among the most effective approaches to date are systems based on lasers \cite{Brunner2013a}, memristors \cite{Tuma2016}, and spin-torque oscillators \cite{Torrejon2017}. In optical implementations of neural networks, inter-neuronal connections rely on various physical mechanisms, including holography \cite{Psaltis1990}, diffraction \cite{Bueno2018, Lin2018}, integrated Mach–Zehnder modulator networks \cite{Shen2017}, wavelength-division multiplexing \cite{Tait2017}, and optical interconnects fabricated using 3D printing technologies \cite{Moughames2020, Dinc2020, Moughames2020a}. In addition, particular attention should be given to Ref.~\cite{McMahon2023}, which reports an optical implementation of neural networks, as well as to studies on hardware-realized networks \cite{Chen2023, Wang2022, MourgiasAlexandris2022, Manuylovich2024, Ma2025}, including optical neural network implementations \cite{Wang2022} and demonstrations of robust, high-speed training of deep networks using coherent silicon photonics \cite{MourgiasAlexandris2022}.

As in any physical system, hardware neural networks are subject to unwanted external perturbations and noise that may contaminate the useful signal. The characteristics of such noise and its impact on signal quality depend on the specific properties of a given experimental setup. It is therefore important to identify which noise levels can significantly affect the overall performance of the network and which can be effectively tolerated without degrading its operation. It should be emphasized that we do not consider noise in the input data, but rather internal noise arising within the network itself. Such noise may originate from imperfections in the input or output channels of individual hardware neurons, as well as from the interconnections between them. As a reference, we adopt the noise characteristics from hardware setup in Ref.~\cite{Brunner2013a}. However, in the present study, a broader range of noise intensities is considered in order to ensure that the results are applicable to a wider class of hardware neural networks.

Our group has previously investigated the effects of noise on trained deep networks \cite{Semenova2022NN}, recurrent networks \cite{Semenova2025EPJ}, and convolutional networks \cite{Kolesnikov2025Chaos}. In addition, based on analytical predictions and numerical simulations, several noise-reduction strategies for hardware neural networks have been proposed \cite{Semenova2022Chaos,Semenova2024Chaos}. Another important result is that introducing internal noise into neurons during training can serve as an effective strategy for constructing robust hardware neural networks \cite{Kolesnikov2026CSF}. The present work represents the final stage in this series of studies on deep neural networks. Its key distinction lies in the comparison of noise introduced before and after the activation function, corresponding to perturbations in the input and output channels of artificial neurons, respectively. We identify a notable effect: sigmoid-type activation functions (including both the sigmoid and hyperbolic tangent) act as efficient nonlinear filters, significantly mitigating the adverse impact of noise. Furthermore, we analyze the applicability of previously proposed noise-reduction strategies \cite{Semenova2022Chaos,Semenova2024Chaos} in the case where noise is present in the input channels of artificial neurons.

\section{System under study}\label{sec:systemUnderStudy}
\subsection{Network}\label{sec:systemUnderStudy:nets}
This article investigates the influence of internal noise on the performance of deep feedforward neural networks. To address this problem, networks of varying depth are considered and trained on the classical task of handwritten digit recognition based on the MNIST dataset \cite{LeCun1998}. Our results indicate that, in general, they do not qualitatively depend on the dataset under consideration. For instance, in our previous work \cite{Kolesnikov2026CSF}, we demonstrated that the effects of noise on both the training and testing processes did not differ qualitatively across the MNIST, Fashion-MNIST, and CIFAR-10 datasets.

Due to the peculiarities of the MNIST dataset and the nature of the classification task, all considered networks have the same number of neurons in the input and output layers. The input layer consists of 784 linear neurons, corresponding to the $28\times 28$ pixel resolution of MNIST images (see Ref.~\ref{sec:systemUnderStudy:training} for details). Since the task is to assign each input image to one of 10 possible classes (digits 0--9), the output layer comprises 10 neurons with a softmax activation function. Consequently, the network’s prediction is determined not by the raw output values themselves, but by the neuron that produces the highest activation in response to a given input image.

\begin{figure}[t]
\includegraphics[width=\linewidth]{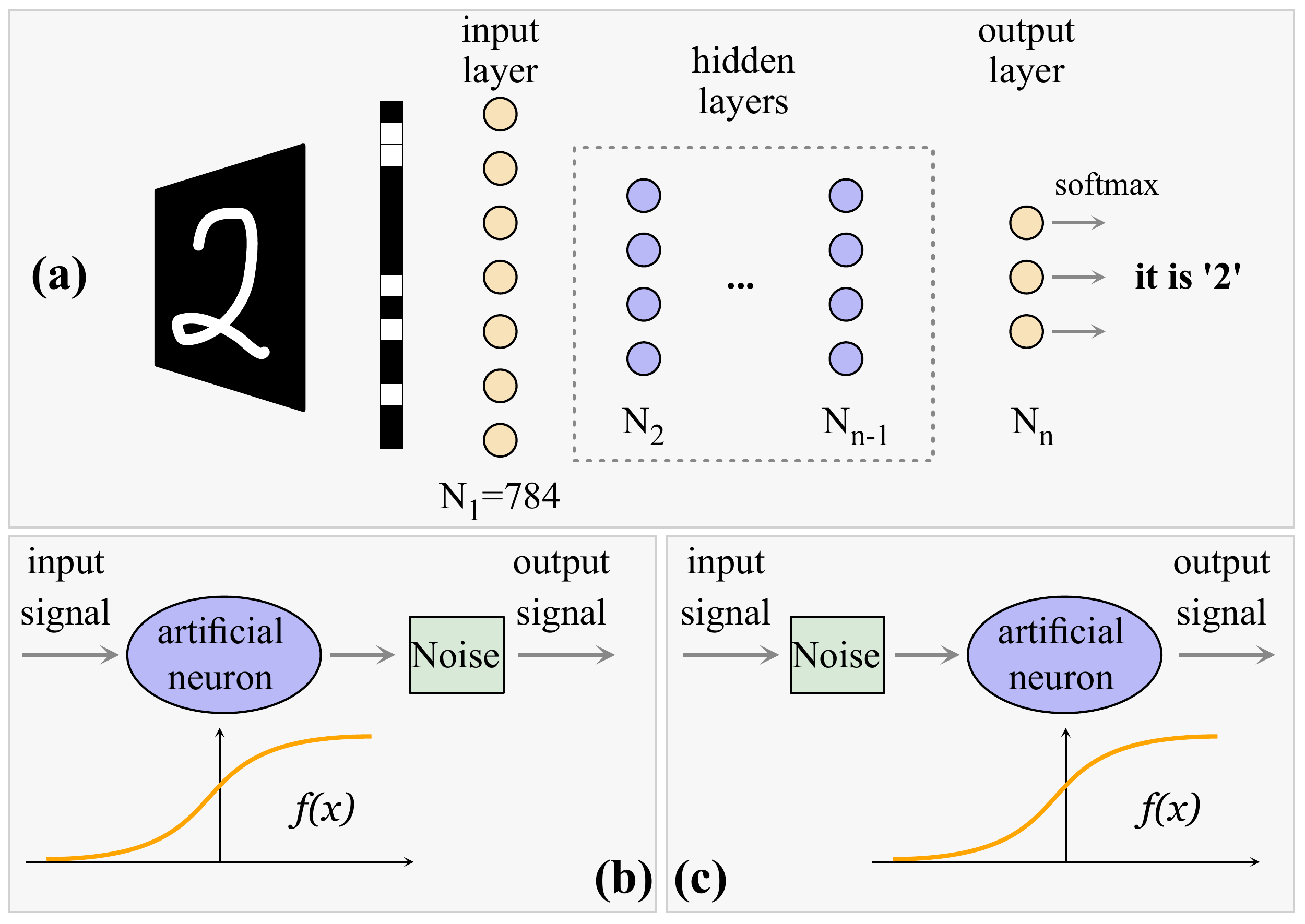}
\caption{\label{fig:scheme}Schematic representation of considered deep neural networks (a) and stages of noise introduction (b),(c).}
\end{figure}

In this paper, we consider feedforward neural networks, in which the signal propagates strictly in one direction from the input layer to the output layer. A schematic representation of the networks under study is shown in Fig.~\ref{fig:scheme}(a). The layers located between the input and output layers are referred to as hidden layers.
In Sections \ref{sec:noiseAfter:depth} and \ref{sec:noiseBefore:depth}, we examine several configurations with different network depths, i.e., varying numbers of hidden layers. In Sections \ref{sec:noiseAfter:number} and \ref{sec:noiseBefore:number}, we investigate different numbers of neurons within the hidden layers and analyze how this affects the accumulation of noise in the overall network output.
Additionally, we study how noise introduced at different stages and within different hidden layers influences the network’s classification accuracy.

In the following, the input signal to the $i$-th neuron in the $n$-th layer is assumed to be given by
\begin{equation}\label{eq:x_n}
\tilde{x}^n_i = \sum\limits_{j=1}^{N_{n-1}} x^{n-1}_j \cdot W^n_{j,i}, \ \ \ \ \tilde{\mathbf{x}}^n = \mathbf{x}^{n-1}\cdot \mathbf{W}^n,
\end{equation}
where $\mathbf{x}^{n-1}$ denotes the vector of output signals from the neurons of the previous layer $(n-1)$, and $N$ represents the number of neurons in the corresponding layer. The matrix $\mathbf{W}^n$ defines the connections between the neurons of layer $n$ and the previous one $n-1$, and has dimensions $N_{n-1}\times N_{n}$. For example, the weight matrix connecting the input layer to a hidden layer consisting of 20 neurons has dimensions $784 \times 20$. 

The output signal of the $i$-th neuron in the $n$-th layer, in the absence of noise, is given by
\begin{equation}\label{eq:y_n}
x^n_i = f(\tilde{x}^n_i) = f\big(\sum\limits_{j=1}^{N_{n-1}} x^{n-1}_j \cdot W^n_{j,i} \big),
\end{equation}
where $f(x) = 1/(1+e^{-x})$ is the activation function.

\subsection{Training}\label{sec:systemUnderStudy:training}

As noted in the previous section, the network is trained using the MNIST dataset \cite{LeCun1998}. This dataset contains 70,000 grayscale images of size $28\times 28$ pixels. Of these, 60,000 images are used for training, and the remaining 10,000 for testing. When working with MNIST, specific conditions are imposed on the input and output layers. The input layer is designed such that each neuron receives the value of the corresponding pixel from the image. Since the images are 28$\times$28 pixels in size, the input layer must consist of $784=28^2$ neurons. For computational convenience, pixel values are normalized by 255, so that the input values fall within the range $[0,1]$. Since the network is trained for a classification task, each input image is assigned to one of 10 classes (digits 0–9). Accordingly, the output layer contains 10 neurons, each corresponding to a specific digit. The network’s classification is determined not by the raw output values of the output neurons themselves, but by identifying the neuron with the highest activation. For example, if an image of the digit 0 is presented to the network, the neuron corresponding to 0 must produce the maximum output. This is implemented using the softmax function.

The networks were trained using the TensorFlow/Keras library \cite{Keras}, employing the Adam optimizer and categorical cross-entropy as the loss function. The achieved accuracy on the training and testing datasets depends strongly on the specific network topology. For instance, the lowest recognition accuracy was observed for a network with a single hidden layer containing 20 neurons, yielding 94.8\% on the training set and 93.9\% on the testing set. In the following experiments, as we vary the number of hidden layers and the number of neurons per layer, the precise recognition accuracy is expected to fluctuate slightly.

\subsection{Noise influence}\label{sec:systemUnderStudy:noise}

This study investigates the effect of white Gaussian noise on the network’s performance. The noise source is modeled as zero-mean, unit-standard-deviation Gaussian noise $\xi$. The noise term $\sqrt{2D}\xi$ was introduced either additively or multiplicatively into the corresponding signal, such that the variance of the noise was defined as twice the noise intensity $2D$. The types of noise and their implementation were inspired by Ref.~\cite{Brunner2013a}, in which a hardware neural network was realized in an optical experiment. However, in the present study, a broader range of noise intensities is considered in order to extend the applicability of the results.

In our previous studies, we considered noise that was introduced into the artificial neurons only after the activation function:
\begin{equation}\label{eq:noise_after}
\begin{array}{c}
y^n_i = x^n_i\cdot(1+\sqrt{2D_M}\xi^M_{n,i}) + \sqrt{2D_A}\xi^A_{n,i} = \\
f\Big(\sum\limits_{j=1}^{N_{n-1}} y^{n-1}_j \cdot W^n_{j,i} \Big) \cdot(1+\sqrt{2D_M}\xi^M_{n,i}) + \sqrt{2D_A}\xi^A_{n,i} .
\end{array}
\end{equation}
Here, the subscript `A' denotes additive noise, while `M' denotes multiplicative noise. It is assumed that the noise sources are uncorrelated, meaning that the noise produces different values for different neurons across layers. These values also vary between different input images. A schematic representation of this method of noise introduction is shown in Fig.~\ref{fig:scheme}(b).

The key distinction of the present work is the comparison of the effects of the stage at which noise is introduced into an artificial neuron. Specifically, noise can be applied either after or before the activation function. Physically, introducing noise after the activation corresponds to distortion in the output channel of a neuron, whereas introducing noise before the activation is equivalent to the presence of noise in the neuron’s input channel. Both scenarios are relevant in the context of hardware neural networks.

In the case where additive or multiplicative noise is applied to the input channel of a neuron, the equations are modified as follows:
\begin{equation}\label{eq:noise_before}
\begin{array}{c}
y^n_i = f\Big(\tilde{x}^n_i\cdot(1+\sqrt{2D_M}\xi^M_{n,i}) + \sqrt{2D_A}\xi^A_{n,i} \Big) = \\
f\Big(\sum\limits_{j=1}^{N_{n-1}} y^{n-1}_j \cdot W^n_{j,i}  \cdot(1+\sqrt{2D_M}\xi^M_{n,i}) + \sqrt{2D_A}\xi^A_{n,i} \Big) .
\end{array}
\end{equation}
A schematic representation of this case is shown in Fig.~\ref{fig:scheme}(c).

\section{Noise after activation function}\label{sec:noiseAfter}
\begin{figure}[h]
\includegraphics[width=\linewidth]{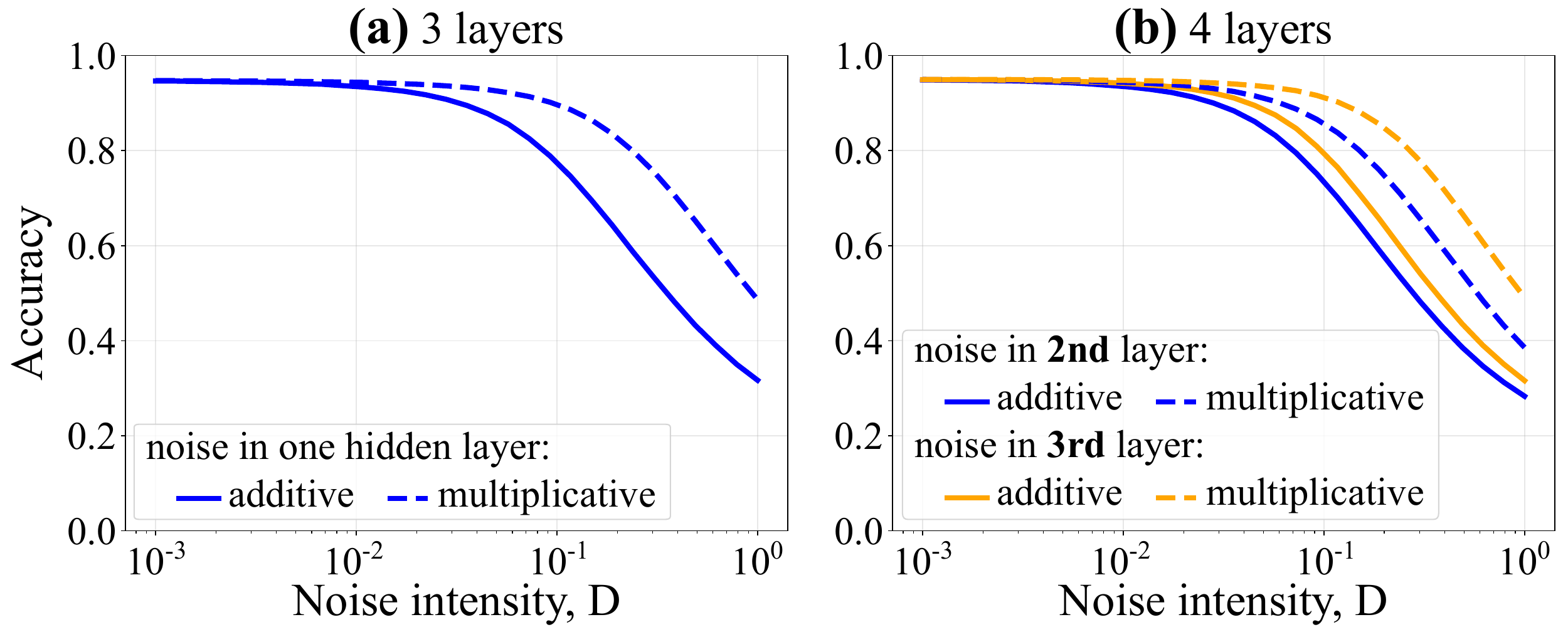}
\caption{\label{fig:after:depth:34} Impact of additive (solid curves) and multiplicative (dashed curves) noise of varying intensities on the accuracy of trained deep neural networks with one hidden layer (a) and two hidden layers (b).}
\end{figure}

We first consider a trained neural network with 20 neurons in the hidden layer. The network was trained using backpropagation with the Adam optimizer, achieving an accuracy of $94.43\% \pm 0.21$ on the training set and $94.17\% \pm 0.23$ on the testing set. The numerical investigation of noise effects on network performance is conducted in a similar manner. Initially, the network is trained using existing Python implementations with TensorFlow and Keras libraries. Then, the trained weights are extracted, and the network is evaluated in a reconstructed form with internal noise introduced as described in Section \ref{sec:systemUnderStudy:noise}. For each noise type and its intensity, the accuracy on testing dataset is calculated (see for example Fig.~\ref{fig:after:depth:34}(a)). Due to the characteristics of high-intensity noise, each point on the plots is averaged over 10 runs on the test dataset.

In this section, noise is introduced into the hidden layer neurons after the activation function, as described in (\ref{eq:noise_after}) and shown in Fig.~\ref{fig:scheme}(b). This approach corresponds to the presence of noise within the artificial neuron or at its output. For a network with a single hidden layer, noise is applied to the second layer. Figure~\ref{fig:after:depth:34}(a) shows the dependence of accuracy on noise intensity $D$ (logarithmic scale on the x-axis). The noise influence was introduced additively (solid blue curve) and multiplicatively (dashed blue curve).

A comparison of the curves in Fig.~\ref{fig:after:depth:34}(a) indicates that additive noise (solid blue line) is more detrimental to network performance than multiplicative noise (dashed blue line). Under additive noise, accuracy remains nearly constant for $D\lesssim 10^{-2}$ but drops sharply to 31.6\% as $D$ increases from $10^{-2}$ to $1$. In contrast, multiplicative noise is less critical: the decline begins only at $D\approx 10^{-1}$, with accuracy remaining at 47.8\% when $D=1$.

\subsection{Network's depth}\label{sec:noiseAfter:depth}

To gain a more comprehensive understanding of the robustness of deep neural networks to noise, we examine the effects of noise introduced in different layers of deeper networks. Adding another hidden layers allows us to investigate how noise accumulates and propagates through the network architecture, as well as how different types of noise (additive and multiplicative) affect the network’s accuracy.

We begin by considering a four-layer network: an input layer, two hidden layers, and an output layer (Fig.~\ref{fig:after:depth:34}(b)). The trained networks achieved an accuracy of $94.46\% \pm 0.22$ on the training set and  $94.28\pm0.2\%$ on the test set. Noise is sequentially introduced into the neurons of the second and third layers, applied after the activation function. This approach allows for a quantitative assessment of how network depth influences robustness under different levels of noise.


Figure~\ref{fig:after:depth:34}(b) presents the curves on a logarithmic scale, which are similar in shape to those obtained for the three-layer network (a). Curves for additive noise are shown as solid lines for noise in the second layer (blue) and for noise in the third layer (orange). Dashed lines of the same colors represent the results for multiplicative noise. As in the three-layer network, multiplicative noise is less critical than additive noise. Moreover, accuracy is lower when noise is added to the second layer compared to the third layer, indicating that the earlier noise enters the network, the more it propagates through subsequent layers, degrading the model’s performance.

In our previous work \cite{Semenova2022NN}, it was shown analytically that noise accumulation depends significantly on the statistical properties of the weight matrices following the noisy layer. Table~\ref{tab:nets:3:4} presents the statistical characteristics of the weight matrices for the considered three- and four-layer networks. Specifically, the mean $\mu$ and the mean square $\eta$ values.

\begin{table}[h]
\caption{\label{tab:nets:3:4} Statistics of connection matrices of ANNs with 3 and 4 layers.}
\begin{ruledtabular}
\begin{tabular}{c|cc|ccc}
      &  \multicolumn{2}{c|}{3 layers} & \multicolumn{3}{c}{4 layers} \\
\hline
Matrix & $\mathbf{W}^2$ & $\mathbf{W}^3$ & $\mathbf{W}^2$ & $\mathbf{W}^3$ & $\mathbf{W}^4$ \\
\hline
Mean value, $\mu(\mathbf{W}^n)$ & 0.0027 & -0.2462 & 0.0116 & 0.0125 & -0.1947 \\
$\Big(\mu(\mathbf{W}^i)\Big)^2$ & 0.0001 & 0.0606 & 0.0001 & 0.0002 & 0.0379 \\
Mean square, $\eta(\mathbf{W}^n)$ & 0.0674 & 1.4845 & 0.0693 & 0.8631 & 1.4139 \\
$N_{n-1}\cdot\eta(\mathbf{W}^n)$ & 52.8416 & 29.6900 & 54.3312 & 17.2620 & 28.2780 \\
\end{tabular}
\end{ruledtabular}
\end{table}

Finally, we consider a deeper network with three hidden layers (five layers in total). As in the previous networks, each hidden layer contains 20 neurons. Noise is introduced separately into each hidden layer. The trained networks achieve an accuracy of $94.03\% \pm 0.44$ on the training set and $93.80\% \pm 0.37$ on the test set.

\begin{figure}[h]
\includegraphics[width=\linewidth]{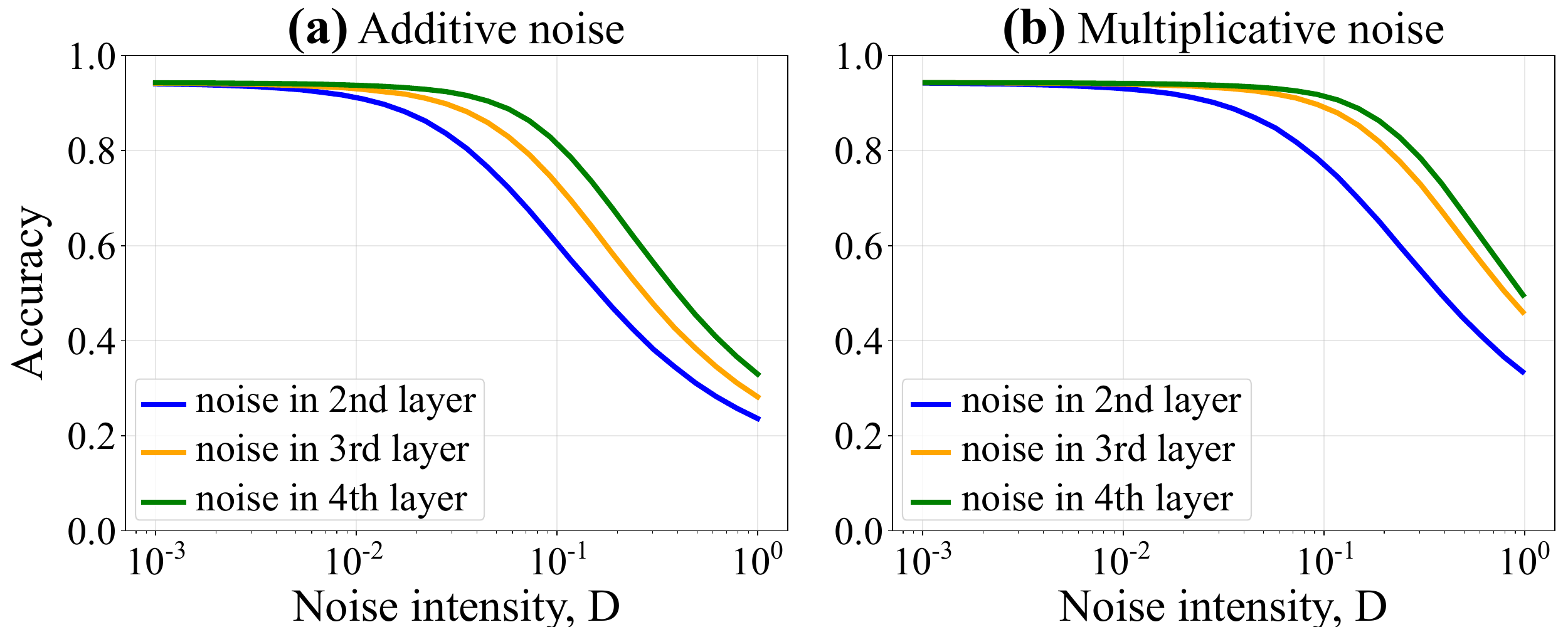}
\caption{\label{fig:after:depth:5} Impact of additive (panel (a)) and multiplicative (panel (b)) noise of different intensities on the accuracy of trained deep neural networks with 5 layers (3 hidden). The noise influences were introduced separately into the 2nd (blue curves), 3rd (orange curves) and 4th layer (green curves).}
\end{figure}

Figure~\ref{fig:after:depth:5} shows the accuracy curves for additive noise (a) and multiplicative noise (b), exhibiting a behaviour similar to that observed in the two-hidden-layer network in Fig.~\ref{fig:after:depth:34}(b). This indicates that, for the given task and architecture, noise introduced at earlier stages propagates more strongly through the network: adding noise to the first hidden layer reduces accuracy more than adding it to the last layer. As noise intensity increases, accuracy declines more sharply, with the network demonstrating greater robustness to multiplicative noise, since for the same intensity, accuracy decreases more significantly under additive noise. Increasing the number of hidden layers further does not fundamentally change this behaviour: noise continues to propagate through the network, earlier layers exhibit lower accuracy, and overall training performance decreases, resulting in a lower final accuracy. As with the previous networks, Table~\ref{tab:nets:5} presents the statistical characteristics of the weight matrices for the network under consideration.

\begin{table}[h]
\caption{\label{tab:nets:5} Statistics of connection matrices of ANN with 5 layers (20-20-20).}
\begin{ruledtabular}
\begin{tabular}{c|cccc}
Matrix & $\mathbf{W}^2$ & $\mathbf{W}^3$ & $\mathbf{W}^4$ & $\mathbf{W}^5$ \\
\hline
Mean value, $\mu(\mathbf{W}^n)$ & 0.0075 & -0.0066 & 0.0374 & -0.1917 \\
$\Big(\mu(\mathbf{W}^i)\Big)^2$ & 0.0001 & 0.00004 & 0.0014 & 0.0367 \\
Mean square, $\eta(\mathbf{W}^n)$ & 0.0575 & 0.6654 & 0.8497 & 1.1893 \\
$N_{n-1}\cdot\eta(\mathbf{W}^n)$ & 45.1027 & 13.3073 & 16.9944 & 23.7856 \\
\end{tabular}
\end{ruledtabular}
\end{table}

Thus, two fundamentally important features of noise accumulation in deep networks have been established. First, multiplicative noise is less detrimental compared to additive noise of the same intensity. Second, the introduction of noise in earlier hidden layers leads to a more pronounced degradation in network accuracy.
It is also important to note that, according to our previous analytical studies of noise accumulation, the process is strongly influenced by the statistical properties of the connection matrices following the noisy layer, in particular by the mean-square value $N_n\eta(\mathbf{W}^{n+1}) = N_n\cdot\frac{1}{N_n N_{n+1}}\sum\limits^{N_n}_{j=1}\sum\limits^{N_{n+1}}_{i=1} W^{n+1}_{j,i} $. For all networks considered, the number of neurons in the hidden layers was fixed, while the mean-square values of the weight matrices increased with depth (see the last row in Tables \ref{tab:nets:3:4} and \ref{tab:nets:5}). For noise accumulation in a 3-layer network, the characteristics of the matrix $\mathbf{W}^3$ are important, for a 4-layer network -- the matrices $\mathbf{W}^3$, $\mathbf{W}^4$, while for a 5-layer network -- the matrices $\mathbf{W}^3$, $\mathbf{W}^4$, $\mathbf{W}^5$. This explains why introducing noise into earlier layers results in a more significant degradation of network performance. If, after each layer, the variance is multiplied by factors such that $N_n\eta(\mathbf{W}^{n+1})>1$ for all subsequent layers, this leads to an overall growth of the variance. However, for networks with the same number of neurons per layer, we were unable to obtain the opposite regime.
 
\subsection{Number of neurons}\label{sec:noiseAfter:number}
We now examine how the number of neurons in the hidden layers affects the network performance. For this purpose, we consider a deep network consisting of five layers: an input layer, three hidden layers, and an output layer. The number of neurons in the input and output layers is fixed, as it is determined by the task and the dataset. Therefore, in this section, we vary only the number of neurons in the hidden layers. In the previous section, a network with a 20-20-20 hidden-layer configuration was analyzed. Here, we consider alternative topologies with 10-10-10, 30-30-30, and 350-250-200 neurons in the hidden layers.

Multiple networks with identical topologies were trained, resulting in essentially identical results across all cases. Therefore, representative results for a single network of each type are presented.

\begin{figure}[h]
\includegraphics[width=\linewidth]{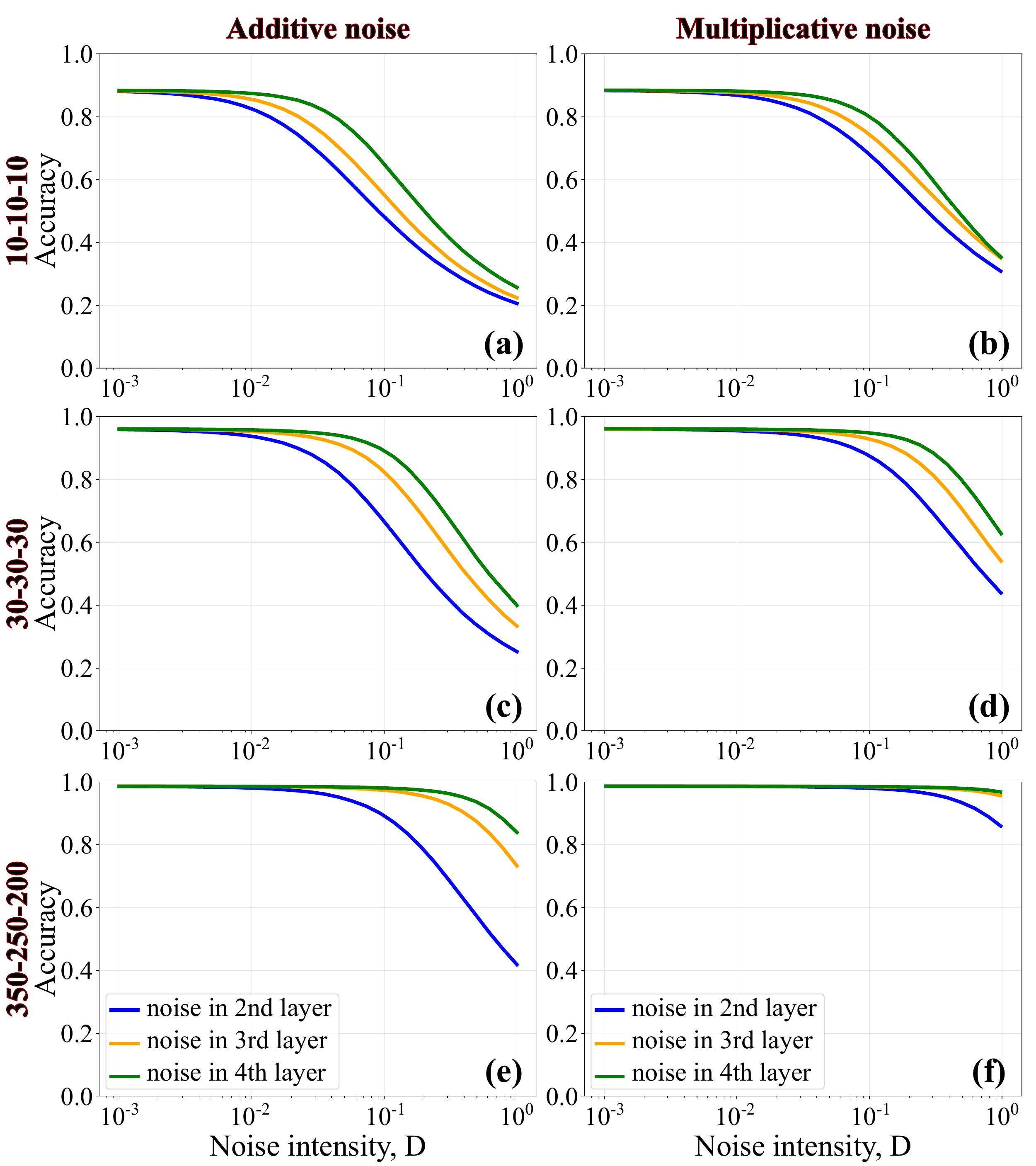}
\caption{\label{fig:after:number:5} Impact of additive (left panels) and multiplicative (right panels) noise of different intensities on the accuracy of trained deep neural networks with 5 layers (3 hidden). The noise influences were introduced separately into the 2nd (blue curves), 3rd (orange curves) and 4th layer (green curves) of trained networks with number of neurons in hidden layers: 10-10-10 (a,b), 30-30-30 (c,d) and 350-250-200 (e,f). }
\end{figure}

Figure~\ref{fig:after:number:5} presents the dependence of network accuracy on noise intensity for additive (a, c, e) and multiplicative (b, d, f) noise. The panels shows results for different network topologies, with colors indicating the specific hidden layer into which the noise was introduced.

For the network with a 10-10-10 hidden-layer configuration, the training accuracy reached $88.26\%\pm 2.04$, while the testing accuracy was $87.6\%\pm 1.04$. The effect of noise on this network is shown in Fig.~\ref{fig:after:number:5}(a,b).

For the network with a 30-30-30 hidden-layer configuration, the training accuracy reached $95.65\%\pm 0.13$, while the testing accuracy was $95.28\%\pm 0.23$. The effect of noise on this network is shown in Fig.~\ref{fig:after:number:5}(c,d).

Empirically, it was observed that the highest accuracy is achieved by networks in which the number of neurons decreases from the input layer toward the output. For example, for a network with a 350-250-200 hidden-layer configuration, the training accuracy reached $98.30\%\pm 0.04$, while the testing accuracy was $97.46\%\pm 0.18$. The effect of noise on this network is shown in Fig.~\ref{fig:after:number:5}(e,f) while the statistics of connection matrices is given in Table~\ref{tab:nets:5_350_250_200}.

\begin{table}[h]
\caption{\label{tab:nets:5_350_250_200} Statistics of connection matrices of ANN with 5 layers (350-250-200).}
\begin{ruledtabular}
\begin{tabular}{c|cccc}
Matrix & $\mathbf{W}^2$ & $\mathbf{W}^3$ & $\mathbf{W}^4$ & $\mathbf{W}^5$ \\
\hline
Mean value, $\mu(\mathbf{W}^n)$ & -0.014734 & -0.012411 & -0.008957 & -0.017416 \\
$\Big(\mu(\mathbf{W}^i)\Big)^2$ & 0.000217 & 0.000154 & 0.000080 & 0.000303 \\
Mean square, $\eta(\mathbf{W}^n)$ & 0.021685 & 0.025420 & 0.020364 & 0.034164 \\
$N_{n-1}\cdot\eta(\mathbf{W}^n)$ & 17.00104 & 8.897 & 5.091 & 6.8328 \\
\end{tabular}
\end{ruledtabular}
\end{table}

Comparing the changes in network accuracy under noisy conditions leads to the same qualitative conclusions as in the previous section. The number of neurons does not affect the fundamental mechanisms of noise accumulation. However, quantitatively, networks with a larger number of neurons in hidden layers exhibit greater robustness to noise. This can be attributed to the averaging of noisy neuron outputs as signals propagate through the connection matrices to subsequent layers. In agreement with the observations for varying network depth in the previous section, additive noise is more detrimental to accuracy than multiplicative noise. Furthermore, introducing noise in earlier layers results in a more significant degradation in performance compared to introducing the same noise in later layers.

\subsection{Noise reduction strategy}\label{sec:noiseAfter:noiseReduction}
In our previous works \cite{Semenova2022Chaos, Semenova2024Chaos}, techniques for reducing noise in deep neural networks were proposed. For the types of noise considered, the pooling technique can be regarded as optimal. The idea of the method is to replace each neuron in a layer with a group of $m$ identical neurons, each receiving the same input signal. The subsequent weight matrix is modified such that the outputs of each group of identical neurons are averaged. Due to this averaging, a substantial reduction in noise can already be achieved for $m=2$, and increasing the group size further improves noise suppression. A key feature of our previous studies \cite{Semenova2022Chaos, Semenova2024Chaos} is that the pooling technique was applied under the assumption that all neurons in the hidden layers are subject to noise. In the present work, we investigate how this technique performs when noise is introduced separately into different layers of the network, with pooling applied sequentially to each layer accordingly.

\begin{figure}[h]
\includegraphics[width=\linewidth]{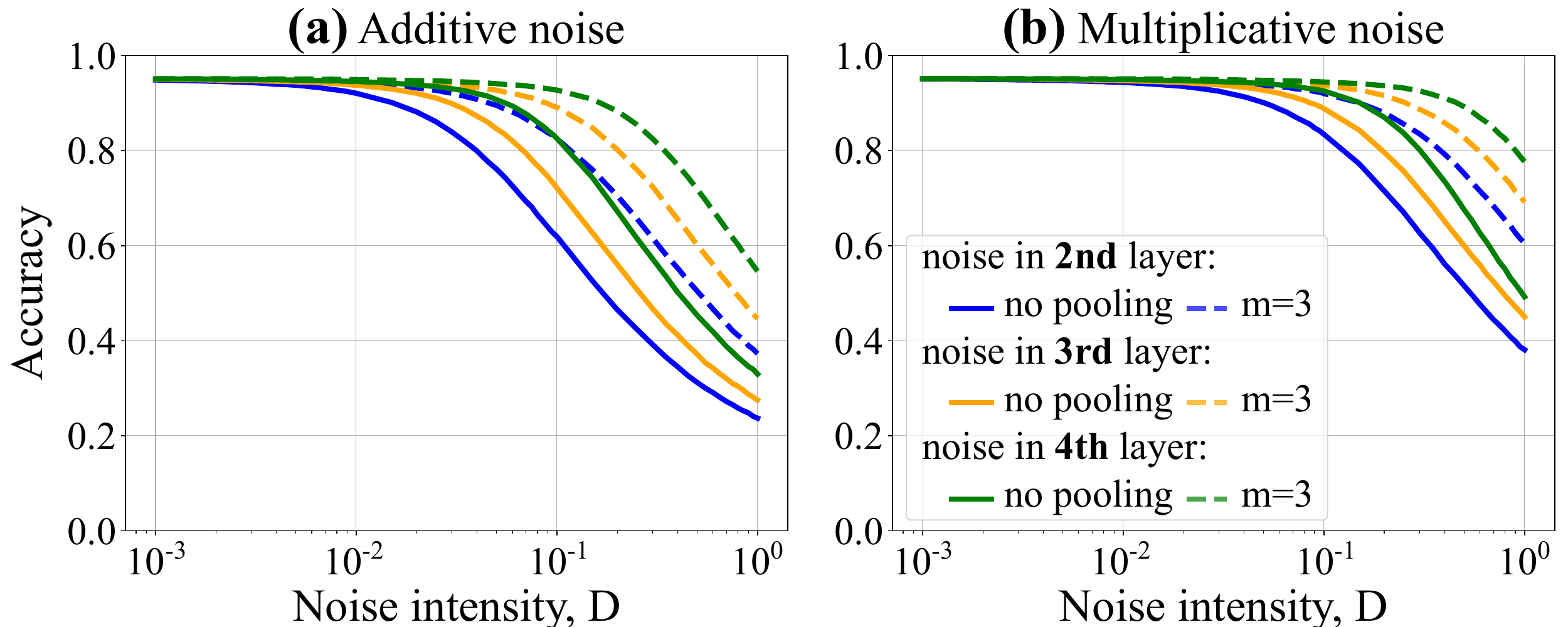}
\caption{\label{fig:after:pooling:5} Noise reduction pooling technique with $m=3$ for networks with additive (a) and multiplicative (b) noise of different intensities \emph{after} activation function. The noise influences were introduced separately into the 2nd (blue curves), 3rd (orange curves) and 4th layer (green curves) of trained networks. Solid lines were obtained for networks without noise reduction, while dashed lines were obtained for pooling with $m=3$.}
\end{figure}

Figure~\ref{fig:after:pooling:5} illustrates the performance of a trained five-layer network with the topology described in the previous section under different intensities of additive (a) and multiplicative (b) noise. Curves of different colors correspond to noise introduced into different layers. Solid lines indicate the accuracy achievable without applying any noise-reduction techniques, while dashed lines show the results obtained when the pooling technique is applied with groups of $m=3$ neurons.

As shown in Ref.\cite{Semenova2022Chaos}, the use of this technique reduces the signal variance by a factor of $m^2$, although its impact on the network’s accuracy is also positive but less pronounced \cite{Semenova2024Chaos}. In the present work, we specifically consider the case $m=3$, which significantly mitigates the influence of noise without eliminating it entirely. From the curves in Fig.~\ref{fig:after:pooling:5}, it can be seen that noise cannot be completely suppressed, and noise introduced in earlier layers has a more critical impact on the network. At the same time, the improvements due to the pooling technique are less noticeable in earlier layers, supporting the hypothesis that noise accumulation is driven by the subsequent weight matrices. This issue can be addressed by applying the pooling technique across all hidden layers, including those that are not directly affected by noise.

\section{Noise before activation function}\label{sec:noiseBefore}
In our previous works \cite{Semenova2022NN, Semenova2025EPJ, Kolesnikov2025Chaos, Semenova2022Chaos, Semenova2024Chaos} and in the previous section, we considered the effect of noise introduced after the activation function, which is analogous to perturbations in the output channel of an artificial neuron. However, it is equally important to examine the impact of noise when it is introduced before the activation function (\ref{eq:noise_before}), as this case accounts for perturbations in the input channel of the neuron or in the connections from preceding neurons.

\begin{figure}[h]
\includegraphics[width=\linewidth]{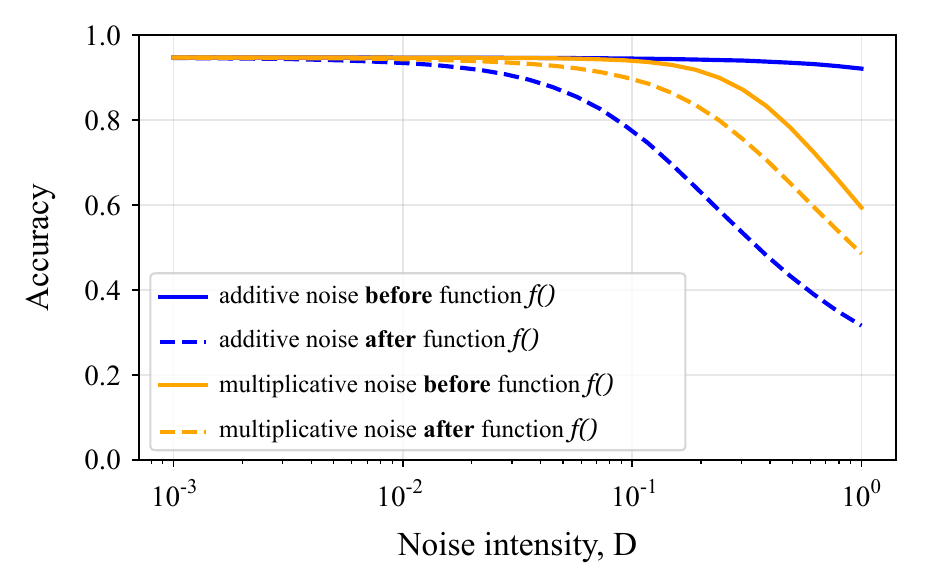}
\caption{\label{fig:before:depth:3} Impact of additive (blue curves) and multiplicative (orange curves) noise of different intensities on the accuracy of trained deep neural networks with one hidden layer of 20 neurons. The noise was introduced before activation functon (solid lines) and after (dashed lines).}
\end{figure}

Analogously to the three-layer network in Fig.~\ref{fig:after:depth:34}(a), we now consider the case where noise is introduced before the activation function. These results are presented in Fig.~\ref{fig:before:depth:3} as solid lines. The network consists of a single hidden layer with 20 neurons. For comparison, the results obtained when noise is introduced after the activation function are also shown by dashed lines of the same colors. The blue curves correspond to additive noise, while the orange curves represent multiplicative noise, illustrating how accuracy varies with increasing noise intensity.

A comparison of the curves in Fig.~\ref{fig:before:depth:3} shows that the activation function acts as an effective filter for noise: the solid curves corresponding to noise introduced before the activation function lie significantly above the corresponding dashed curves obtained when noise is introduced after the activation function. Moreover, the results indicate that additive noise is filtered more effectively in this setting.

\subsection{Network's depth}\label{sec:noiseBefore:depth}
\begin{figure}[h]
\includegraphics[width=\linewidth]{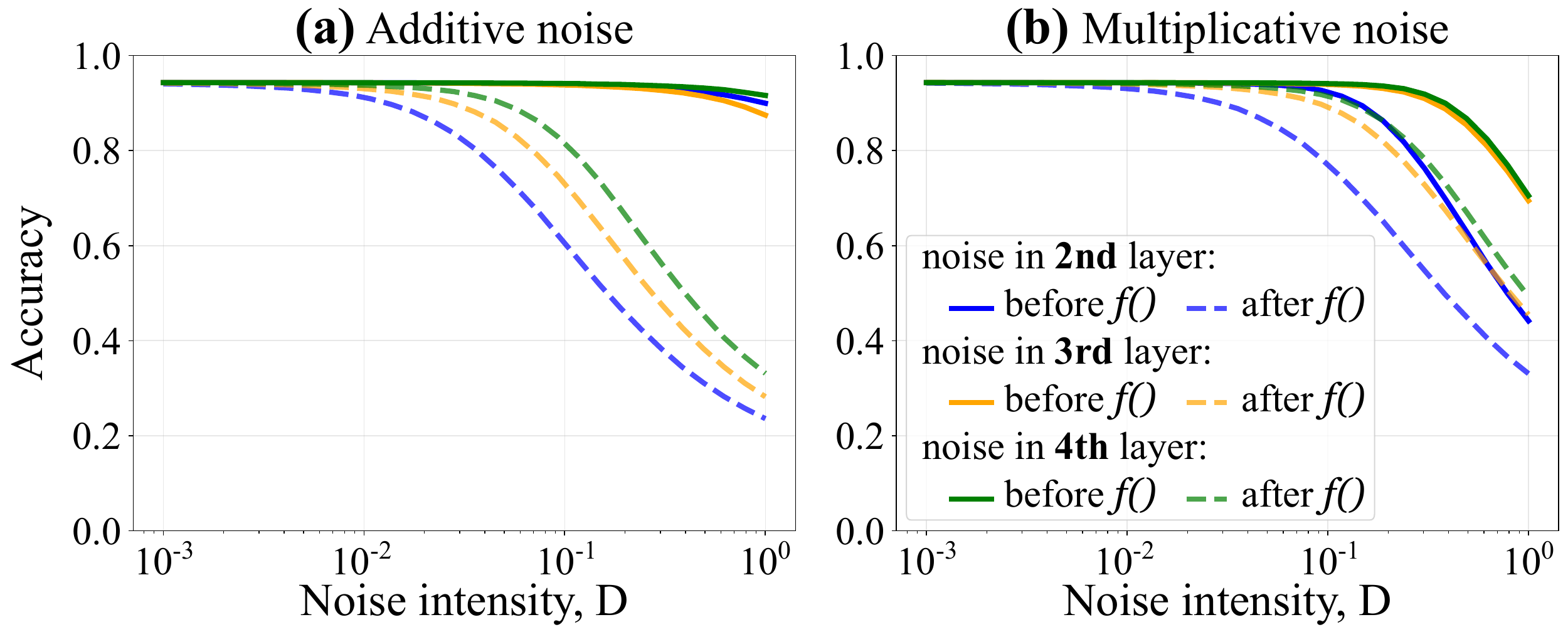}
\caption{\label{fig:before:depth:5} Impact of additive (a) and multiplicative (b) noise of different intensities on the accuracy of trained 5-layer neural networks with 20-20-20 hidden layers. The noise was introduced before activation function (solid lines) and after (dashed lines) into the 2nd (blue), 3rd (orange) and 3rd (green) layers separately.}
\end{figure}

We now consider a deeper network consisting of three hidden layers with 20 neurons each. This network was previously analyzed in Section \ref{sec:noiseAfter:depth} in Fig.~\ref{fig:after:depth:5}. Here, we compare how the accuracy of this network changes when noise is introduced before the activation function. The corresponding results are given in Fig.~\ref{fig:before:depth:5}.

Similar results were obtained for the four-layer network and for deeper architectures. Regardless of the layer into which noise is introduced, the accuracy of networks with noise applied before the activation function is significantly higher than that of networks where noise is introduced after the activation function. This effect is particularly pronounced for additive noise. As for the relative ordering of the curves when noise is introduced into deeper layers, it is not uniquely determined by the statistical properties of the weight matrices. Due to the nonlinearity of the activation function, the behaviour may vary considerably. For example, in Fig.~\ref{fig:before:depth:5}(b) corresponding to multiplicative noise, the curves preserve the same ordering both before and after the activation function. In contrast, in Fig.~\ref{fig:before:depth:5}(a), this ordering changes when additive noise is introduced before the activation function.

\subsection{Number of neurons}\label{sec:noiseBefore:number}
\begin{figure}[h]
\includegraphics[width=\linewidth]{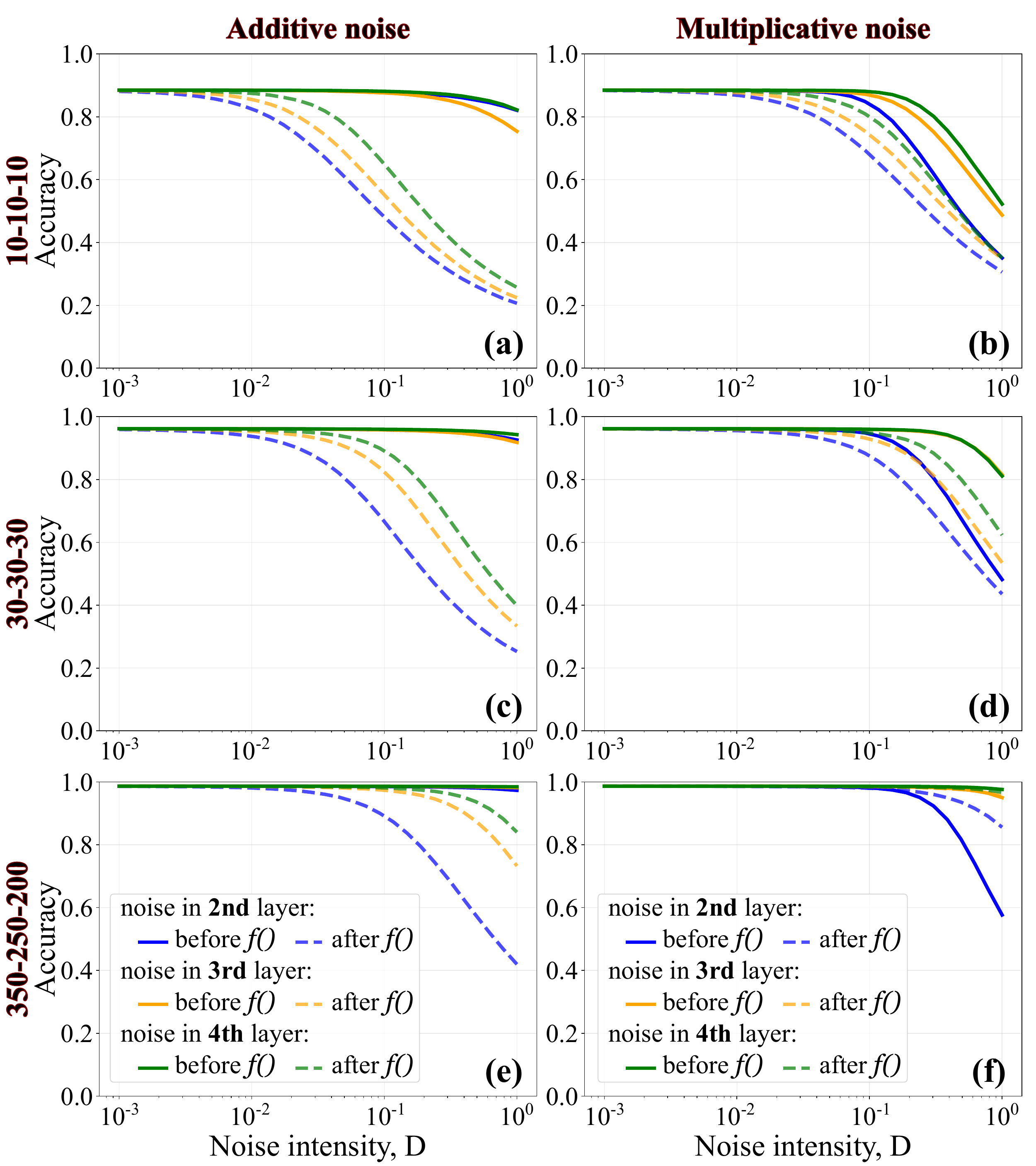}
\caption{\label{fig:before:number:5} Impact of additive (left panels) and multiplicative (right panels) noise of different intensities on the accuracy of trained deep neural networks with 5 layers (3 hidden). The noise influences were introduced separately into the 2nd (blue curves), 3rd (orange curves) and 4th layer (green curves) of trained networks with number of neurons in hidden layers: 10-10-10 (a,b), 30-30-30 (c,d) and 350-250-200 (e,f). The results for networks with noise before activation function are shown by solid lines, while dashed lines correspond to the noise after activation function. }
\end{figure}

In this section, we consider five-layer networks similar to those examined in Section \ref{sec:noiseAfter:number}, with hidden-layer topologies of 10-10-10, 30-30-30, and 350-250-200. In Fig.~\ref{fig:before:number:5}, solid lines correspond to networks with noise introduced before the activation function, while dashed lines represent the case where noise is introduced after the activation function.

From the plots, it is evident that for networks with noise introduced before the activation function, all the results and conclusions presented above remain independent of the number of neurons in the hidden layers.

\subsection{Noise reduction strategy}\label{sec:noiseBefore:noiseReduction}

\begin{figure}[h]
\includegraphics[width=\linewidth]{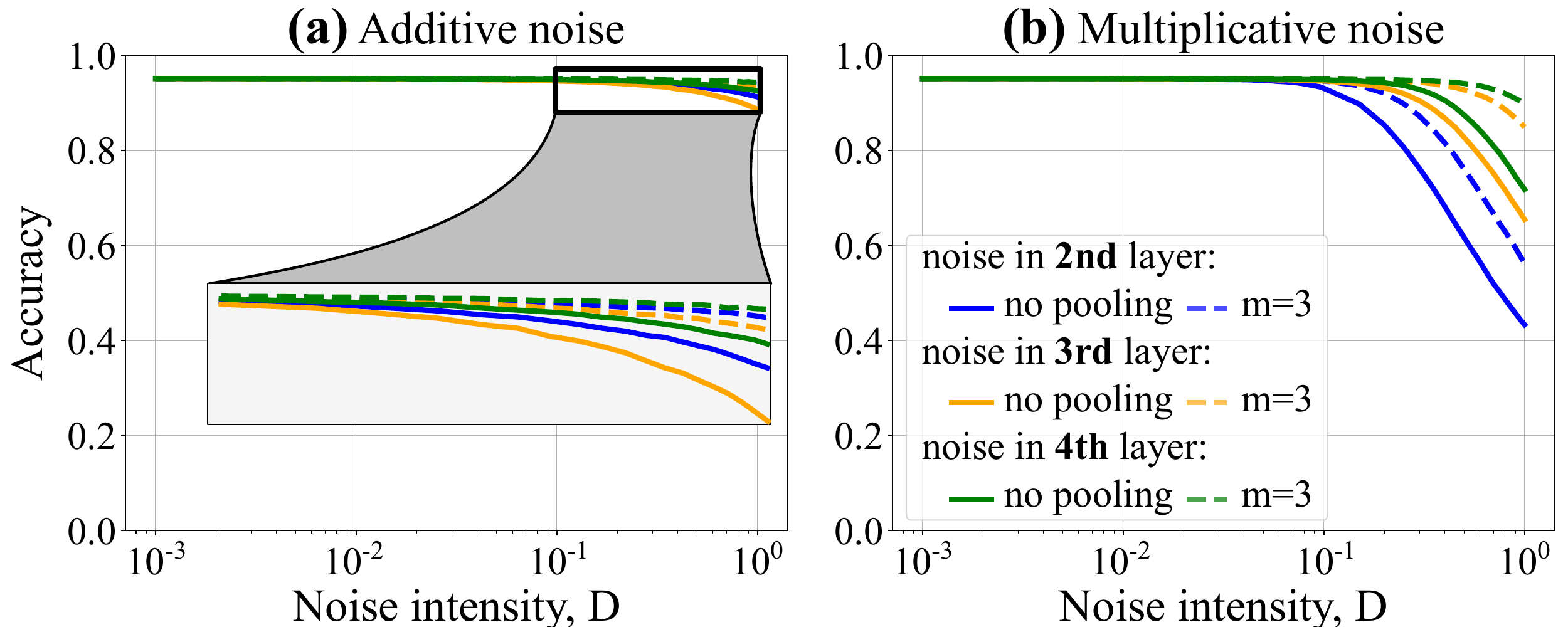}
\caption{\label{fig:before:pooling:5} Noise reduction pooling technique with $m=3$ for networks with additive (a) and multiplicative (b) noise of different intensities \emph{before} activation function. The noise influences were introduced separately into the 2nd (blue curves), 3rd (orange curves) and 4th layer (green curves) of trained networks. Solid lines were obtained for networks without noise reduction, while dashed lines were obtained for pooling with $m=3$. }
\end{figure}

We now examine the applicability of the noise-reduction method based on pooling, which has demonstrated strong performance for noise introduced after the activation function (see Section~\ref{sec:noiseAfter:noiseReduction}). The results for networks with noise introduced before the activation function are given in Fig.~\ref{fig:before:pooling:5}.

When pooling is applied before the activation function (Fig.~\ref{fig:before:pooling:5}), the accuracy improves substantially, consistent with the behaviour observed for the after-function case (Fig.~\ref{fig:after:pooling:5}). However, given that additive noise is already effectively suppressed by the activation function, the inclusion of pooling even with $m=3$ leads to a great enhancement in accuracy, approaching nearly the same level as that of the noise-free network.

\section{Conclusion}\label{sec:conclu}

The activation function acts as a significant filter for noise. For the same trained networks, the accuracy obtained with noise introduced before the activation function is substantially higher than that achieved for the same noise intensities when the noise is applied after the activation function. Moreover, the results indicate that additive noise is filtered more effectively in this case.

For networks with noise introduced before the activation function, multiplicative noise is more critical, whereas for networks with noise introduced after the activation function, additive noise has a more pronounced negative effect.

\textbf{Noise after the activation function.} Two fundamentally important features of noise accumulation in the network were established. First, multiplicative noise is less critical than additive noise of the same intensity. Second, introducing noise in earlier hidden layers leads to a degradation in network accuracy. It is also important to note that, according to our previous analytical studies of noise accumulation, the process is strongly influenced by the statistical properties of the weight matrices following the noisy neuron layer, in particular the mean-square value: $N_n\eta(\mathbf{W}^{n+1}) = N_n\cdot\frac{1}{N_n N_{n+1}}\sum\limits^{N_n}_{j=1}\sum\limits^{N_{n+1}}_{i=1} W^{n+1}_{j,i} $. For all networks considered with different number of layers and neurons, the mean-square value of the weight matrices increased with network depth (see the last column in Tables \ref{tab:nets:3:4} and \ref{tab:nets:5}). This explains why introducing noise into earlier layers leads to a more pronounced degradation in network performance. If, after each layer, the variance is multiplied by subsequent factors such that $N_n\eta(\mathbf{W}^{n+1})>1$, this results in an overall growth of the variance. However, we were not able to obtain the opposite regime for various network topologies. In our previous paper \cite{Kolesnikov2026CSF}, it was shown that only introduction of noise during training signification change the statistics of connection matrices to the opposite case.

\textbf{Noise before the activation function.} The activation function acts as a significant filter for noise. For the same trained networks, the accuracy obtained with noise introduced before the activation function is substantially higher than that achieved for the same noise intensities when the noise is applied after the activation function. Moreover, the results indicate that additive noise is filtered more effectively in this case. At the same time, the identification of which layer is more vulnerable to noise is no longer as clearly determined by the statistical properties of the weight matrices as in the case of noise introduced after the activation function. This behaviour is attributed to the strong nonlinearity of the activation function. For networks with noise introduced before the activation function, all the results and conclusions presented above remain independent of the number of neurons in the hidden layers.

\textbf{Noise reduction.} In this work, we investigated the applicability of a noise-reduction method based on pooling with duplication of noisy neurons \cite{Semenova2022Chaos, Semenova2024Chaos}. Here we show that the pooling technique consistently improves the performance of noisy networks regardless the moment of introducing the noise (before/after the activation function and the layer). At the same time, in certain cases, the statistical properties of the subsequent connection matrices may lead to amplification of noise. In particular, it is possible that the combined effect of noise introduction and pooling in the first hidden layer may be effectively nullified by the following connections. To mitigate this issue, we recommend applying pooling in all subsequent layers, regardless of the specific layer in which noise is introduced.

However, this limitation primarily pertains to the case where noise is applied after the activation function. Since the activation function itself already serves as an effective noise filter, when noise is introduced before the activation function, it is generally sufficient to apply pooling only in the layer where the noise is present.

\begin{acknowledgments}
This work was supported by the Russian Science Foundation (project No. 25-72-10055). 
\end{acknowledgments}

\section*{Data Availability Statement}
The data that support the findings of this study are available from the corresponding author upon reasonable request.

\bibliography{bibliography}

\end{document}